\RequirePackage{pdfmanagement-testphase}
\DocumentMetadata{}

\documentclass[10pt,twocolumn,letterpaper]{article}

\usepackage[pagenumbers]{wacv} 

\usepackage{graphicx}
\usepackage{amsmath}
\usepackage{amssymb}
\usepackage{booktabs}

\usepackage[pagebackref,breaklinks,colorlinks]{hyperref}
\usepackage{cuted}
\usepackage{capt-of}
\usepackage{url}
\usepackage{breakurl}
\usepackage{multirow}

\usepackage[accsupp]{axessibility}

\newcommand{\ykA}[1]{#1}
\newcommand{\yoA}[1]{#1}

\usepackage[capitalize]{cleveref}
\crefname{section}{Sec.}{Secs.}
\Crefname{section}{Section}{Sections}
\Crefname{table}{Table}{Tables}
\crefname{table}{Tab.}{Tabs.}

\def\wacvPaperID{1605} 
\def\confName{WACV}
\def\confYear{2024}

\begin{document}

\title{DiffBody: Diffusion-based Pose and Shape Editing of Human Images}

\author{Yuta Okuyama, Yuki Endo, and Yoshihiro Kanamori\\
Unversity of Tsukuba\\
{\tt\small okuyama.yuta.sw@alumni.tsukuba.ac.jp, {}\{endo, kanamori\}@cs.tsukuba.ac.jp}
}



\maketitle

\thispagestyle{empty}
\begin{strip}\centering
\includegraphics[width=\textwidth]{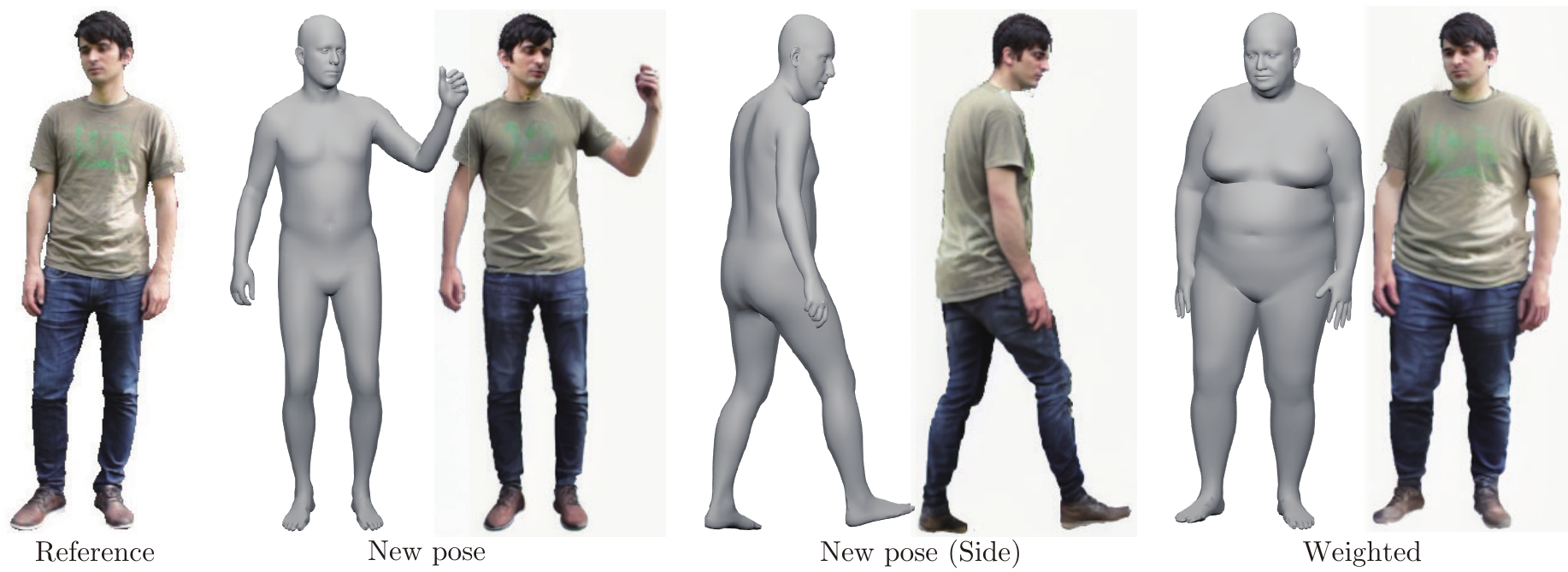}
\captionof{figure}{Results of editing the pose and body shape of \ykA{a} fullbody human image using our method. For various pose and body shape inputs, our method can generate realistic human images, while preserving the clothing textures and facial identity of reference person images.
\label{fig:teaser.}}
\end{strip}

\begin{abstract}
\ykA{Pose and body shape editing in a human image has received increasing attention. However, current methods often struggle with dataset biases and deteriorate realism and the person's identity when users make large edits. We propose a one-shot approach that enables large edits with identity preservation. To enable large edits, we fit a 3D body model, project the input image onto the 3D model, and change the body's pose and shape. Because this initial textured body model has artifacts due to occlusion and the inaccurate body shape, the rendered image undergoes a diffusion-based refinement, in which strong noise destroys body structure and identity whereas insufficient noise does not help. We thus propose an iterative refinement with weak noise, applied first for the whole body and then for the face. We further enhance the realism by fine-tuning text embeddings via self-supervised learning. Our quantitative and qualitative evaluations demonstrate that our method outperforms other existing methods across various datasets. \url{https://github.com/yutaokuyama/DiffBody}}
\end{abstract}

 
\section{\ykA{Introduction}}
\ykA{Editing the pose and body shape of a human image is a task to change the orientation, position, and slenderness/fatness of the subject's limbs and torso, and outputs a realistic image of the same person.} This task has been actively studied due to its \ykA{potential applications, such as visual simulation after dieting and efficient fashion photo shooting.}

\begin{figure}[t]
\centering
 \includegraphics[width=\linewidth]{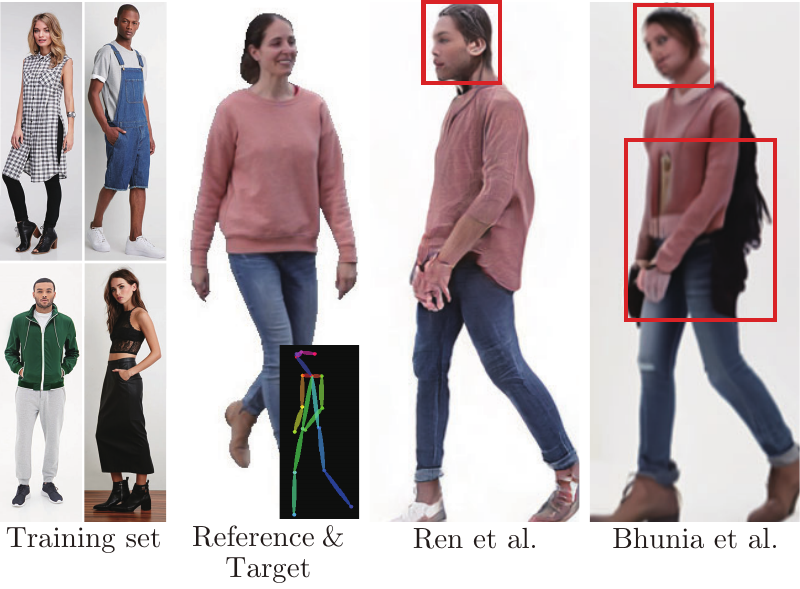}
 \caption{
 Limitations of the state-of-the-art methods for human pose editing by Ren et al.~\cite{NTED} and Bhunia et al.~\cite{PIDM}. These methods often struggle with identity preservation for input images not observed in a test set corresponding to a training set (see the red boxes). 
 }
 \label{fig:OOD_failed}
\end{figure}

\ykA{Recent techniques for editing pose and shape in a human image can be broadly categorized into two approaches, i.e., 1) the approach based on image warping and generative adversarial networks (GANs) and 2) the diffusion-based approach~\cite{PIDM}. The former approach utilizes image warping and well preserves the person's identity, but it often causes artifacts with large edits (see Figures~\ref{fig:qualitative} and \ref{fig:BodyShapeEdit}). The latter approach yields high-quality output images with diverse poses and shapes thanks to the diffusion models. However, it often struggles with identity preservation; the output human figures tend to have different clothes or faces from those of the input image (see Figure~\ref{fig:OOD_failed}).}

\begin{figure}[t]
\centering
     \includegraphics[width=\linewidth]{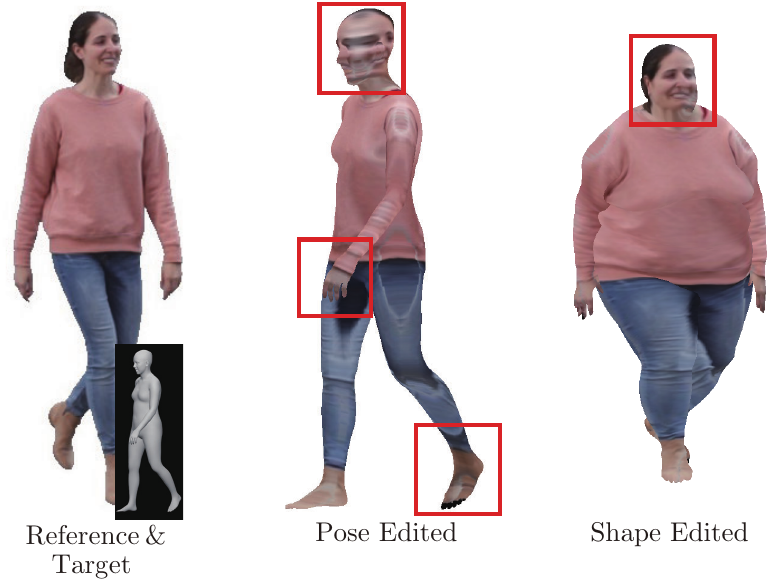}
     \caption{
     Problem of person image projection onto 3D parametric body models. As shown in the red boxes, the initial textured model often has artifacts when its pose and shape are edited. 
     }
 \label{fig:Failed_project}
\end{figure}

\begin{figure}[t]
\centering
 \includegraphics[width=\linewidth]{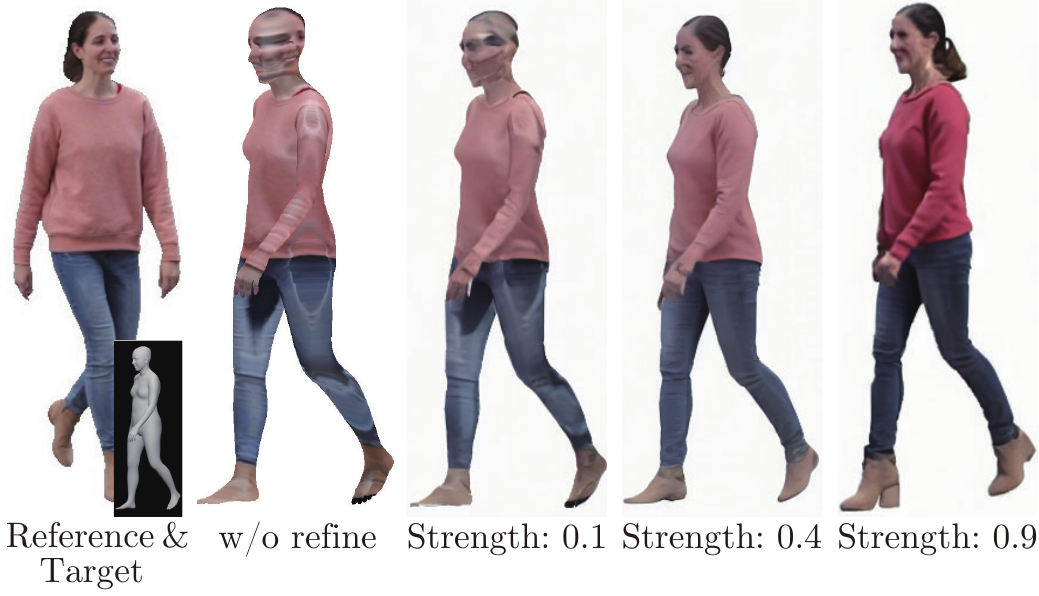}
 \caption{
 Refinement results of a textured model using a diffusion model with different noise strengths. Insufficient noise strength does not refine the visual artifacts, whereas excessively strong noise will significantly alter the person's identity. 
 }
 \label{fig:Failed_Diffusion}
\end{figure}

\ykA{In this paper, we propose a method to enable large edits for various poses and shapes in a single human image with identity preservation. Our large edit for poses and shapes is driven by a 3D parametric body model, followed by refinement using a diffusion model. Namely, we fit a 3D parametric body model~\cite{SMPL-X:2019} to the reference human image, project the reference image onto the 3D model, and change the pose and shape parameters to obtain a textured human model with a new pose and shape. This initial textured model often has artifacts (see Figure~\ref{fig:Failed_project}) because it exhibits occluded regions that were not visible in the reference image, and the target body shape of the texture projection is often inaccurate. We thus refine the rendered human image using an image-to-image translation technique with a diffusion model~\cite{meng2021sdedit}. However, a critical pitfall lies in the difficulty of the noise strength control in the diffusion model; insufficient noise strength does not refine the visual artifacts, whereas excessively strong noise will significantly alter the person's identity (see Figure~\ref{fig:Failed_Diffusion}).}

\ykA{To improve the visual artifacts while avoiding significant structural changes, we propose an iterative refinement with weak noise.}
Furthermore, to enhance the quality of refinements, we update text embeddings used for network conditioning during iterative refinements through self-supervised learning. 
Building upon these refinement processes, \ykA{we adopt} a staged pipeline \ykA{in which we first refine} the fullbody image \ykA{and then locally refine facial features to enhance} realism. 
We quantitatively and qualitatively evaluate our method, demonstrating superior results compared to existing methods across various datasets.


\section{Related work}
\paragraph{Image-to-image translation.} To synthesize human images according to specified poses, an early attempt based on image-to-image translation adopted a generative adversarial network (GAN) conditioned on keypoints~\cite{ma2017pose}. Another attempt used a U-Net architecture combined with a variational autoencoder (VAE)~\cite{esser2018variational}. However, these approaches cannot handle large deformation because they struggle to handle feature misalignment between input and output images with different poses. Some methods enable more flexible pose editing by using flow fields~\cite{Ren_2020_CVPR,li2019dense,ma2021fda,jain2023vgflow}, 2D UV maps of human bodies (i.e., DensePose)~\cite{albahar2021pose,Denseposetransfer}, or human parsing maps~\cite{zhang2021pise,zhou2022cross,lv2021learning}. In addition, the attention mechanism~\cite{attnAYN} further improved the quality of generated images by efficiently and extensively capturing image features~\cite{zhu2019progressive,NTED,zhang_2022_CVPR,PoNA,tang2020xinggan}. Recently, PIDM\cite{PIDM} used a diffusion model for pose editing and achieved state-of-the-art results thanks to its capability of image generation. 
These existing methods yield satisfactory results for a test set corresponding to a training set. However, for diverse human images outside the dataset, they often cause discrepancies between a generated person and a reference person in facial identity and clothing textures, resulting in limiting their applicability to in-the-wild data.
In contrast, our method performs well for diverse inputs by refining textured 3D human models using general text-to-image diffusion models finetuned for a reference person image. 
\paragraph{Image warping.}
Several pose editing methods perform image warping based on 3D human models to improve generalizability for diverse human images~\cite{liu2021liquid,PGHA,svitov2023dinar}. 
These methods first fit a 3D human model~\cite{SMPL-X:2019} to a reference person image. Then, the reference person image is projected onto the fitted 3D human model, whose pose is edited to obtain a final image with the desired pose. These methods inpaint invisible regions in reference images by using flow fields~\cite{liu2021liquid}, GANs~\cite{PGHA}, or diffusion models~\cite{svitov2023dinar}. However, these methods still struggle to handle invisible regions and sometimes yield highly distorted textures. 
Instead of directly inpainting invisible regions, our method extends an image-to-image translation technique using a diffusion model~\cite{meng2021sdedit} to refine projected textures. Our iterative refinement approach enables us to generate more plausible textures in invisible regions. 
\paragraph{Text-to-image translation.}
There have also been approaches utilizing large-scale language models and diffusion models for synthesizing person images from text and pose information~\cite{T2I, zhang2023adding}. Combining these methods with DreamBooth~\cite{ruiz2022dreambooth}, which finetunes diffusion models with several reference images, enables pose editing of a specific person in a reference image. The benefits of large-scale language models and diffusion models allow us to generate realistic human images with diverse poses. However, preserving the facial identity and clothing textures of the reference images remains challenging in this approach.
\paragraph{Body shape editing.}
Zhou et al.~\cite{zhou2010parametric} proposed the first method for body shape editing using 3D human models. In this method, the user interactively fits a \ykA{3D parametric body} model, SCAPE~\cite{anguelov2005scape}, to an input fullbody image. Then, after editing the body shape of the 3D model by specifying height and weight parameters, the user can obtain output images deformed via image warping based on the 3D model. MovieReshape~\cite{zhou2010parametric} automated the fitting of SCAPE models by taking a video as input. FBBR~\cite{FlowBodyReshape} is a method that directly applies image warping based on 2D flow field to an input human image. The user can specify a positive or negative value indicating the amount of deformation, and the CNN estimates a flow field for the input reference person image. However, it yields significant distortion when extensively editing body shapes. 
In our method, we perform image deformation based on 3D human models, while locally refining the invisible and distorted regions of the deformed images using diffusion models. This allows large deformation with less distortion. 

\begin{figure}[t]
\centering
 \includegraphics[width=\linewidth]{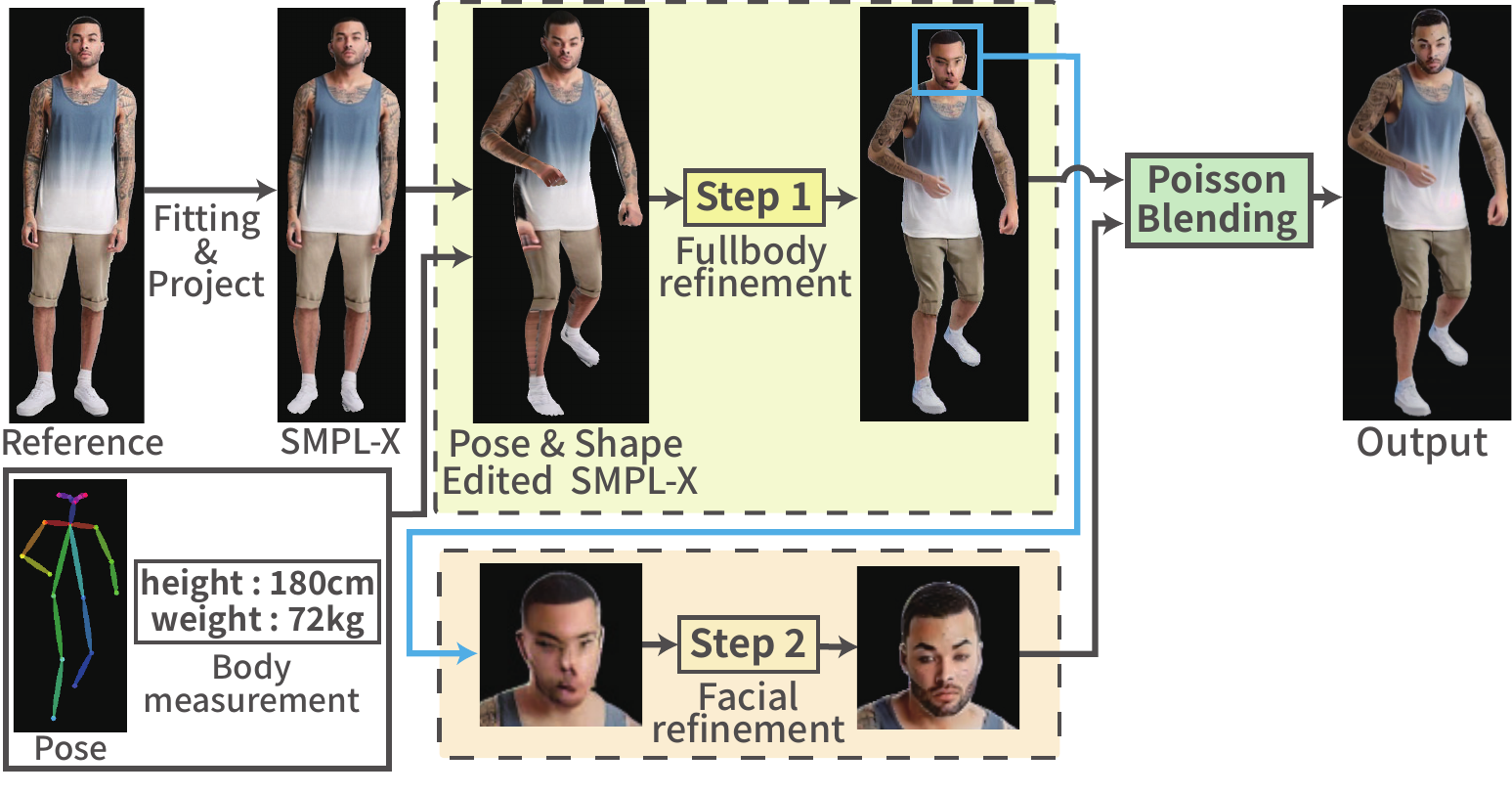}
 \caption{Overview of \ykA{our} method. Our method first computes a textured SMPL-X model from a reference person image. The SMPL-X model is then deformed according to given body pose, height, and weight parameters. To compensate for occluded and distorted textures resulting from texture projection, our method performs step-by-step refinement for the rendered image of the SMPL-X model using diffusion models.\label{fig:NetworkOverview}}
\end{figure}
\section{Method}
Figure~\ref{fig:NetworkOverview} shows an overview of our method. We first reconstruct a 3D human model from a reference image and then improve its appearance via diffusion-based refinement. 
The reconstruction procedure of 3D human models is as follows. We first fit SMPL-X~\cite{SMPL-X:2019}, which is a \ykA{3D} parametric \ykA{body} model, to a reference image. Next, we obtain a texture of the SMPL-X model by projecting the reference image. We then specify body shape parameters (height and weight) and keypoints (joint positions) to manipulate the pose and body shape of the SMPL-X model. Finally, we obtain a coarse human image by rendering the textured SMPL-X model. 

Although the rendered image of a textured SMPL-X can reflect a specified pose and body shape (see Figure~\ref{fig:Failed_project}), it contains visual artifacts due to invisible areas and texture distortion. Therefore, we aim to reduce such artifacts by leveraging diffusion models, which are equipped with highly-expressive generative capability and local editability. We use the pre-trained latent diffusion model (LDM)~\cite{LatentDiffusion} as a backbone of our method. To faithfully preserve a person's identity, we use DreamBooth~\cite{ruiz2022dreambooth}. Specifically, we finetune the LDM using a body image and a face image cropped from a reference image. As a textual input of DreamBooth, we use a prompt containing a special token ``sks" associated with the reference person (see Figure~\ref{fig:NetworkOverview}). We also condition the LDM on keypoints of joint positions using T2I-Adapter~\cite{T2I} to reflect pose information effectively. 

Using the finetuned diffusion model conditioned on a text prompt and keypoints, we refine a textured SMPL-X image. To do so, inspired by SDEdit~\cite{meng2021sdedit}, we perform image-to-image translation that can modify fine details of an input image while preserving its coarse structure via forward (noising) and reverse (denoising) processes. However, simply using SDEdit unnecessarily changes textures in visible areas while not sufficiently modifying visual artifacts in invisible and distorted regions. This is because textures in each area should be modified with different noise levels, but SD\ykA{E}dit uniformly adds noise to an entire image and performs denoising. To address this issue, we propose two-stage refinement steps, which modify each part with different noise levels. The first step is for refining fullbody textures in invisible areas, whereas the second step is for facial distorted textures. We describe these steps in the following sections. 



\begin{figure}[t]
\centering
 \includegraphics[width=\linewidth]{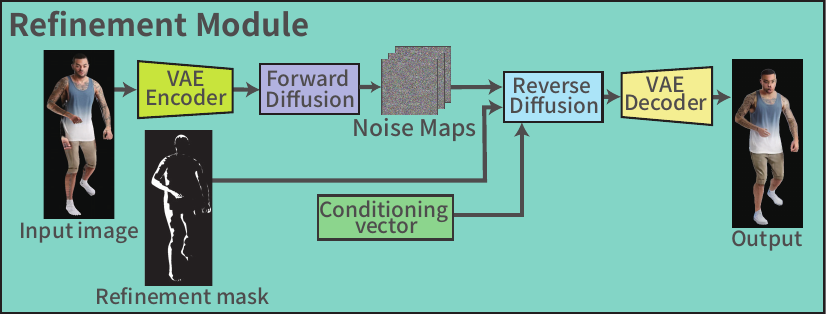}
 \caption{Overview of our refinement module. This module is based on an image-to-image translation technique using a diffusion model~\cite{meng2021sdedit}. It first adds noise to the latent map computed from an input image using the VAE. Then, the diffusion model denoises the noised maps using a refinement mask and conditioning vector extracted from a prompt and keypoints. The final output image is obtained from the denoised latent map using the VAE. 
 }
 \label{fig:Refine}
\end{figure}

\subsection{Step~1: Fullbody refinement}
\label{sec:step1}
In \ykA{Step~1}, we introduce a refinement module (Figure~\ref{fig:Refine}) for correcting unnatural textures in invisible areas caused by the projection of reference images. The refinement module takes as input an image to be modified, a refinement mask for invisible areas, a conditioning vector extracted from a prompt and keypoints. The input image is converted into a latent feature map using the VAE encoder of the LDM. Inspired by Blended Diffusion~\cite{Avrahami_2022_CVPR}, we perform denoising in the refinement mask to modify only invisible areas. 
Specifically, we first add noise to the latent feature map through the forward process and store the noise map at each time step. Then, during the reverse process, we perform denoising while replacing the values outside the refinement mask with the noise map stored at the corresponding time step. However, in this process, excessively strong noise unnecessarily modifies human coarse structure, whereas weak noise cannot improve fine texture details sufficiently. In addition, the appropriate noise strength depends on input images, but adjusting individual noise strengths is cumbersome.


\begin{figure}[t]
\centering
 \includegraphics[width=\linewidth]{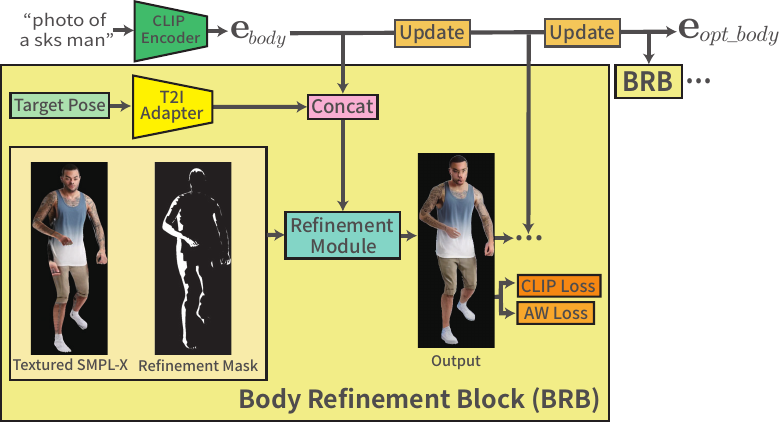}
 \caption{Overview of fullbody refinement in Step~1. The body refinement block (BRB) iteratively refines the invisible area of a textured SMPL-X model using our refinement module. At each refinement iteration, it also updates the text embedding $\mathbf{e}_{body}$ using the CLIP and \yoA{AW} losses. The process of BRB is also repeated while reinitializing the refined image to make the update of the text embedding stable. }
 \label{fig:NetworkStep1}
\end{figure}

\paragraph{Iterative refinement.} To adequately modify fine texture details while preserving coarse structure, our method iteratively refines images via multiple reverse processes with weak noise. As illustrated in Figure~\ref{fig:NetworkStep1}, we iterate the process that adds weak noise to an input image and performs denoising using the refinement module. During the iteration, we need to determine how many iterations are the best to obtain a high-quality output. This optimal iteration number is determined based on loss functions for evaluating the output image. We first use the \yoA{Adaptive Wing (AW) loss~\cite{wingloss}} between joint heatmaps estimated by OpenPose~\cite{OpenPose} for the output image and the rendered SMPL-X image. In addition, we use CLIP similarity~\cite{CLIP} between the output and reference images for each part~\cite{elicit} based on SMPL-X labeling. The final output is obtained when the sum of these loss functions become minimum during fixed iterations. 

\paragraph{Text embedding optimization.} In parallel with iterative refinement, we also aim to further improve the visual quality of the final output by optimizing the text embedding $\mathbf{e}_{body}$ conditioned on the network. At each iteration, we update the text embedding $\mathbf{e}_{body}$ via backpropagation of the above loss functions. 

\paragraph{Input reinitialization.} However, we often observed cases where the refined image becomes unnatural during iterative refinement before the text embedding optimization converges. As a result, the text embedding may also converge to an undesirable solution. To address this issue, we reinitialize the input image with the initial image every several iterations in iterative refinement. This approach makes text embedding optimization more stable and allows us to obtain higher-quality results, as shown in the ablation study in Section~\ref{sec:abs}. 

While this fullbody refinement process can improve the invisible areas of rendered SMPL-X images, unnatural distortions are still observed in face textures, which are further refined in \ykA{Step~2}. 

\begin{figure}[t]
\centering
 \includegraphics[width=\linewidth]{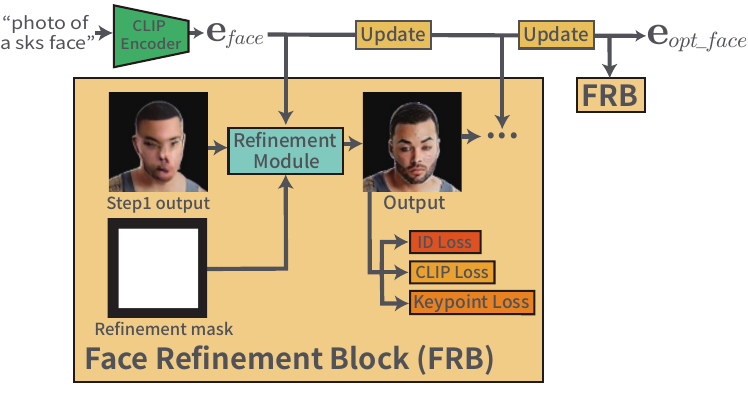}
 \caption{Overview of facial refinement in Step~2. The face refinement block (FRB) refines distorted facial textures in the output of Step~1. Similar to Step~1, we perform iterative refinement using our refinement module. The text embedding $\textbf{e}_{face}$ is also updated according to the ID, CLIP, and \yoA{Keypoint} losses. }
 \label{fig:NetworkStep2}
\end{figure}

\subsection{Step~2: Facial refinement}
Figure~\ref{fig:NetworkStep2} shows an overview of \ykA{Step~2}. 
According to region labels obtained by SMPL-X~\cite{SMPL-X:2019}, we automatically crop a facial region from the image refined in \ykA{Step~1}. After resizing the cropped image to $512 \times 512$, our refinement module iteratively updates the image and text embedding $\mathbf{e}_{face}$ in a similar procedure to \ykA{Step~1}. 
To seamlessly merge face and body images after refinement, we keep the outer region of the face image unchanged by assigning 0 to the outer region of the refinement mask and 1 to the inner region. To optimize the text embedding $\mathbf{e}_{face}$, we use the identity loss by MagFace~\cite{meng2021magface}, 
keypoint loss defined by MSE loss between keypoints of output and rendered SMPL-X images estimated using RetinaFace~\cite{retinaface},
and CLIP similarity~\cite{CLIP}. We obtain a final output in \ykA{Step~2} when the sum of these loss functions shows a minimum value during fixed iterations. The final result is generated by merging the output images in \ykA{Steps~1 and 2} via Poisson blending~\cite{PoissonBlending}.

\section{Experiments}
\paragraph{Datasets.} We used various datasets such as DeepFashion~\cite{DeepFashion}, MonoPerfCap~\cite{MonoPerfCap}, Everybody Dance Now \ykA{(EDN)}~\cite{everybodyDanceNow}, You\ykA{T}ube 18 Dancers~\cite{lee2019metapix}, and EHF~\cite{SMPL-X:2019}. From DeepFashion, we extracted 80 images containing fullbody humans and used 26 images as reference person images and 54 images as target poses. For the other datasets containing videos of different people, we used a single frame of each video as a reference and about 35 frames of each video as targets. In total, we used 51 reference images and 963 target images.

\paragraph{Evaluation metrics.}
We evaluated our method using the metrics including SSIM, PSNR, LPIPS~\cite{lpips}, FID~\cite{fid}, and the \yoA{AW} and ID losses. For the \yoA{AW} loss, we employed the L2 loss between heatmaps estimated from generated and target images using OpenPose~\cite{OpenPose}. For the ID loss, we used the cosine similarity between facial features obtained using MagFace~\cite{meng2021magface}.\\
\textbf{Implementation details.}
We implemented our method using Python and PyTorch and ran our program on NVIDIA RTX A6000. The image size used in the experiments was $512 \times 512$. In our method, we finetuned a pre-trained Stable Diffusion (v1.4)~\cite{LatentDiffusion} 
with \ykA{a single reference image} for each person
using DreamBooth~\cite{ruiz2022dreambooth}. During the finetuning process, we used the AdamW optimizer with a learning rate of $1.0 \times 10^{-6}$. For inference, we used 30\% noise for image refinement in \ykA{Steps~1 and 2} and performed 100 iterations of refinement. For input reinitialization, which aims to make text embedding optimization stable, we reinitialize the input image every 5 iterations in iterative refinement. For text embedding optimization, we used the Adam optimizer~\cite{kingma2014adam} and the cosine annealing scheduler~\cite{loshchilov2016sgdr} with warmup, with minimum and maximum learning rates set to $4.0 \times 10^{-4}$ and $5.0 \times 10^{-4}$, respectively. We used PyMaF~\cite{pymaf2021} to estimate SMPL-X parameters. Our method took about 20 minutes for each finetuning and inference. The resulting images shown in our paper are trimmed to save space.
See our supplementary material for more implementation details and additional results.

\begin{table}[]
\centering
\caption{Quantitative comparison on the Everybody Dance Now (EDN)~\cite{everybodyDanceNow}, 
EHF~\cite{SMPL-X:2019}, MonoPerfCap~\cite{MonoPerfCap}, You\ykA{T}ube 18 Dancers (Y18D)~\cite{lee2019metapix}, DeepFashion~\cite{DeepFashion} datasets, and average scores across all datasets.}
\label{table:Quantitative_all_dataset}
\resizebox{\columnwidth}{!}{%
\begin{tabular}{llcccccc}
\hline
 & \multicolumn{1}{l|}{} & PSNR$\uparrow$ & SSIM$\uparrow$ & LPIPS$\downarrow$ & FID$\downarrow$ & ID$\downarrow$ & \yoA{AW}$\downarrow$ \\ \hline \hline
\multirow{7}{*}{\rotatebox{90}{EDN~\cite{everybodyDanceNow}}} 
& \multicolumn{1}{l|}{LWG~\cite{liu2021liquid}} & 17.000 & 0.751 & 0.251 & 48.322 & 0.352 & 7.70 \\
 & \multicolumn{1}{l|}{PGHA~\cite{PGHA}} & 16.219 & 0.745 & 0.304 & 90.952 & 0.408 & 6.22 \\
 & \multicolumn{1}{l|}{DINAR~\cite{svitov2023dinar}} & 15.398 & 0.725 & 0.276 & 52.515 & 0.308 & 6.78 \\
 & \multicolumn{1}{l|}{NTED~\cite{NTED}} & 16.153 & 0.742 & 0.292 & 90.843 & 0.480 & 5.92 \\
 & \multicolumn{1}{l|}{PIDM~\cite{PIDM}} & 14.928 & 0.680 & 0.320 & 108.288 & 0.487 & 8.76 \\
 & \multicolumn{1}{l|}{T2IA~\cite{T2I}} & 5.027 & 0.400 & 0.627 & 229.689 & 0.454 & 3.90 \\
 & \multicolumn{1}{l|}{Ours} & \textbf{18.574} & \textbf{0.793} & \textbf{0.212} & \textbf{44.801} & \textbf{0.276} & \textbf{3.30} \\ \hline
\multirow{7}{*}{\rotatebox{90}{EHF~\cite{SMPL-X:2019}}} & \multicolumn{1}{l|}{LWG~\cite{liu2021liquid}} & 18.994 & 0.788 & 0.232 & 68.922 & 0.210 & 3.20 \\
 & \multicolumn{1}{l|}{PGHA~\cite{PGHA}} & 17.763 & 0.764 & 0.289 & 124.662 & 0.389 & 2.75 \\
 & \multicolumn{1}{l|}{DINAR~\cite{svitov2023dinar}} & 17.332 & 0.744 & 0.286 & 87.227 & 0.333 & 2.89 \\
 & \multicolumn{1}{l|}{NTED~\cite{NTED}} & 17.564 & 0.751 & 0.323 & 285.534 & 0.473 & 2.31 \\
 & \multicolumn{1}{l|}{PIDM~\cite{PIDM}} & 15.125 & 0.681 & 0.379 & 279.378 & 0.490 & 3.28 \\
 & \multicolumn{1}{l|}{T2IA~\cite{T2I}} & 6.062 & 0.546 & 0.567 & 238.12 & 0.382 & 2.93 \\
 & \multicolumn{1}{l|}{Ours} & \textbf{20.356} & \textbf{0.815} & \textbf{0.216} & \textbf{58.887} & \textbf{0.180} & \textbf{1.62} \\ \hline
\multirow{7}{*}{\rotatebox{90}{MonoPerfCap~\cite{MonoPerfCap}}} & \multicolumn{1}{l|}{LWG~\cite{liu2021liquid}} & 19.133 & 0.696 & 0.269 & \textbf{41.500} & 0.268 & 4.17 \\
 & \multicolumn{1}{l|}{PGHA~\cite{PGHA}} & 17.849 & 0.670 & 0.339 & 88.296 & 0.400 & 4.13 \\
  & \multicolumn{1}{l|}{DINAR~\cite{svitov2023dinar}} & 17.104 & 0.651 & 0.313 & 53.149 & 0.278 & 6.68 \\
 & \multicolumn{1}{l|}{NTED~\cite{NTED}} & 18.064 & 0.662 & 0.318 & 80.670 & 0.476 & 2.23 \\
 & \multicolumn{1}{l|}{PIDM~\cite{PIDM}} & 16.547 & 0.649 & 0.350 & 92.661 & 0.476 & 4.78 \\
& \multicolumn{1}{l|}{T2IA~\cite{T2I}} & 5.133 & 0.382 & 0.621 & 219.755 & 0.424 & 3.10 \\
 & \multicolumn{1}{l|}{Ours} &  \textbf{19.720} & \textbf{0.716} & \textbf{0.242} & 46.432 & \textbf{0.175} & \textbf{1.82} \\ \hline
\multirow{7}{*}{\rotatebox{90}{Y18D~\cite{lee2019metapix}}} & \multicolumn{1}{l|}{LWG~\cite{liu2021liquid}} & 17.000 & 0.701 & 0.285 & 52.927 & 0.328 & 4.30 \\
 & \multicolumn{1}{l|}{PGHA~\cite{PGHA}} & 16.583 & 0.689 & 0.328 & 107.667 & 0.393 & 4.65 \\
  & \multicolumn{1}{l|}{DINAR~\cite{svitov2023dinar}} & 15.796 & 0.672 & 0.311 & 65.441 & 0.326 & 6,52 \\
 & \multicolumn{1}{l|}{NTED~\cite{NTED}} & 16.651 & 0.683 & 0.324 & 91.852 & 0.485 & 2.70 \\
 & \multicolumn{1}{l|}{PIDM~\cite{PIDM}} & 15.676 & 0.659 & 0.341 & 102.664 & 0.459 & 4.57 \\
& \multicolumn{1}{l|}{T2IA~\cite{T2I}} & 5.057 & 0.391 & 0.644 & 213.486 & 0.422 & 4.07 \\
 & \multicolumn{1}{l|}{Ours} &  \textbf{18.048} & \textbf{0.731} & \textbf{0.254} & \textbf{52.170} & \textbf{0.283} & \textbf{2.54} \\ \hline
\multirow{7}{*}{\rotatebox{90}{DeepFashion~\cite{DeepFashion}}} & \multicolumn{1}{l|}{LWG~\cite{liu2021liquid}} & 15.652 & 0.633 & 0.405 & 123.687 & 0.403 & 14.22 \\
 & \multicolumn{1}{l|}{PGHA~\cite{PGHA}} & 14.939 & 0.621 & 0.450 & 180.599 & 0.533 & 11.29 \\
  & \multicolumn{1}{l|}{DINAR~\cite{svitov2023dinar}} & 14.975 & 0.640 & 0.419 & 115.243 & 0.345 & 11.04 \\
 & \multicolumn{1}{l|}{NTED~\cite{NTED}} & \textbf{16.846} & \textbf{0.665} & \textbf{0.353} & \textbf{76.561} & 0.322 & \textbf{3.39} \\
 & \multicolumn{1}{l|}{PIDM~\cite{PIDM}} & 15.854 & \textbf{0.665} & 0.383 & 78.285 & 0.407 & 5.58 \\
& \multicolumn{1}{l|}{T2IA~\cite{T2I}} & 5.049 & 0.380 & 0.634 & 225.615 & 0.517 & 8.22 \\
 & \multicolumn{1}{l|}{Ours} &  15.738 & 0.642 & 0.376 & 98.572 & \textbf{0.296} & 7.13 \\ \hline
\multirow{7}{*}{\rotatebox{90}{Average}} & \multicolumn{1}{l|}{LWG~\cite{liu2021liquid}} & 17.556 & 0.714 & 0.289 & 67.072 & 0.312 & 6.72 \\
 & \multicolumn{1}{l|}{PGHA~\cite{PGHA}} & 16.671 & 0.698 & 0.342 & 118.435 & 0.425 & 5.81 \\
  & \multicolumn{1}{l|}{DINAR~\cite{svitov2023dinar}} & 16.121 & 0.686 & 0.321 & 74.715 & 0.318 & 6.78 \\
 & \multicolumn{1}{l|}{NTED~\cite{NTED}} & 17.064 & 0.701 & 0.322 & 125.362 & 0.447 & 3.31 \\
 & \multicolumn{1}{l|}{PIDM~\cite{PIDM}} & 15.626 & 0.667 & 0.351 & 132.255 & 0.464 & 5.39 \\
& \multicolumn{1}{l|}{T2IA~\cite{T2I}} & 5.266 & 0.420 & 0.619 & 225.333 & 0.440 & 4.44 \\
 & \multicolumn{1}{l|}{Ours} &  \textbf{18.487} & \textbf{0.739} & \textbf{0.260} & \textbf{60.171} & \textbf{0.242} & \textbf{3.28}\\
 \hline
\end{tabular}%
}
\end{table}
\begin{figure*}[t]
\centering
 \includegraphics[width=\linewidth]{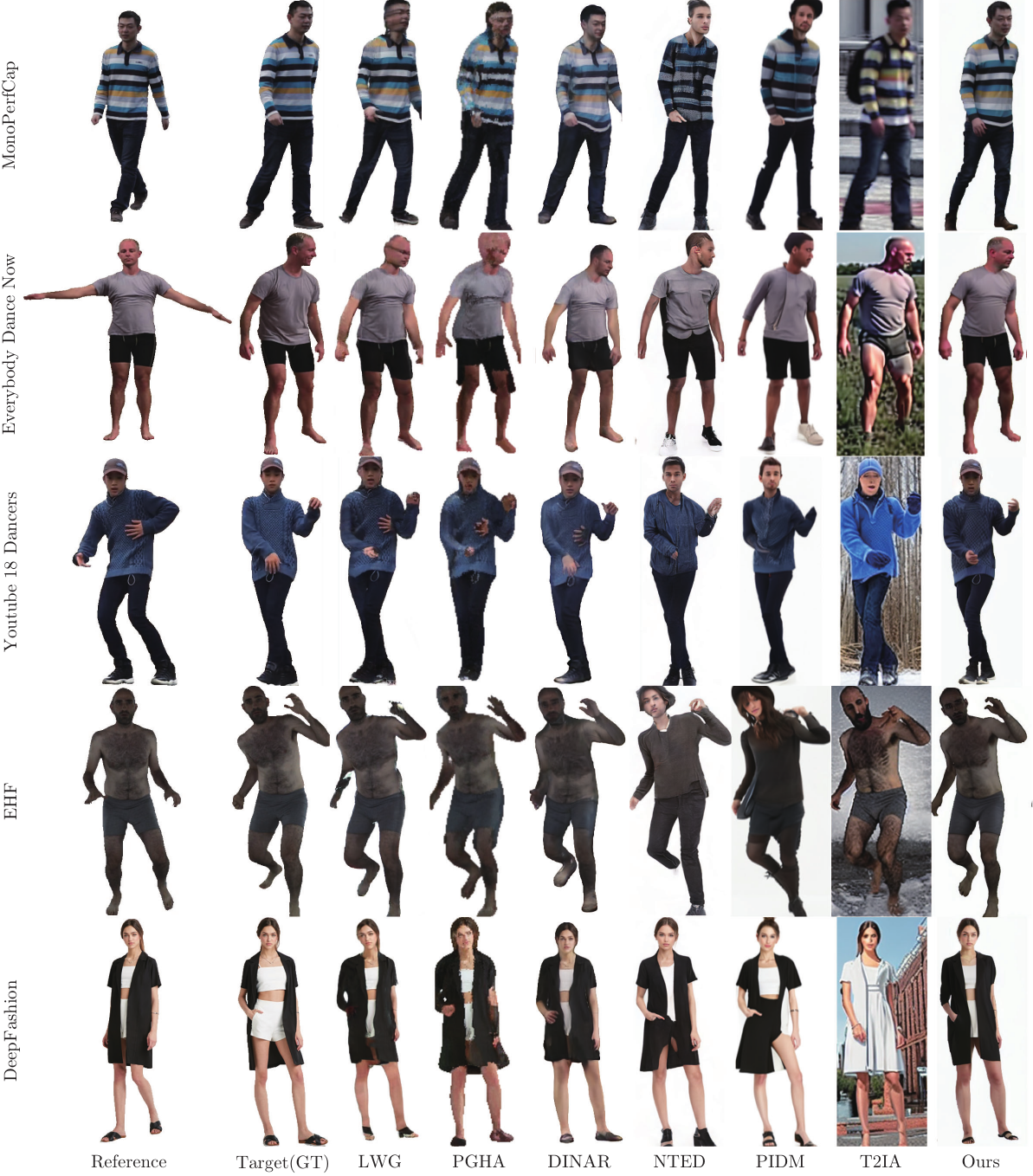}
 \caption{Qualitative comparison of editing the reference images to the target poses with our method and the existing methods (LWG~\cite{liu2021liquid}, PGHA~\cite{PGHA}, DINAR~\cite{svitov2023dinar}, NTED~\cite{NTED}, PIDM~\cite{PIDM}, and T2IA~\cite{T2I}) on each dataset. 
 }
 \label{fig:qualitative}
\end{figure*}


\label{section:result}

\subsection{Evaluation of pose editing}
\paragraph{Quantitative comparison. }Regarding pose editing, we compared our method with the existing methods including LWG~\cite{liu2021liquid}, PGHA~\cite{PGHA}, NTED~\cite{NTED}, PIDM~\cite{PIDM}, T2IA~\cite{T2I}, and DINAR~\cite{svitov2023dinar}. For T2IA, we finetuned the pre-trained diffusion model equipped with the T2I adapter using DreamBooth for each reference person image. For LWG, PGHA, and DINAR, we used official models trained on the iPER~\cite{liu2021liquid}, 3D People~\cite{pumarola20193dpeople}, or Texel~\cite{Texel_2023} dataset. For the remaining methods, we used official models trained on the DeepFashion dataset~\cite{DeepFashion}. Table~\ref{table:Quantitative_all_dataset} shows the quantitative comparison on various datasets. The bold font indicates the best score for each metric on each dataset. As shown in the results, our method outperforms the existing methods in almost all metrics on the four datasets (i.e., Everybody Dance Now, EHF, MonoPerfCap, and You\ykA{T}ube 18 Dancers). In the results on DeepFashion, although NTED performs the best, our method also outperforms the methods not trained on the DeepFashion dataset. Furthermore, our method shows the best average scores across all datasets. These findings suggest that our method works across multiple datasets.
\paragraph{Qualitative comparison. }Figure~\ref{fig:qualitative} shows the qualitative comparison of pose editing on various datasets. In the results of the first and second rows, the warping-based methods such as LWG~\cite{liu2021liquid}, PGHA~\cite{PGHA}, and DINAR~\cite{svitov2023dinar} yield stretched and unnatural textures in the invisible regions of the reference images. Moreover, in the third row, we can see that these methods burn the hand texture into the torso region. The image-to-image translation methods such as NTED~\cite{NTED} and PIDM~\cite{PIDM} show good results on the DeepFashion dataset (the fifth row). However, they struggle to preserve clothing textures and facial identity on other datasets. The text-to-image approach, T2IA~\cite{T2I}, also suffers from this problem and even generates different backgrounds from the reference images. Our method, on the other hand, consistently produces satisfactory results on all of the datasets. 
Our method successfully achieves a wide range of pose editing for a variety of person images, which is difficult to achieve with the existing methods. 

\begin{figure}[t]
\centering
 \includegraphics[width=0.9\linewidth]{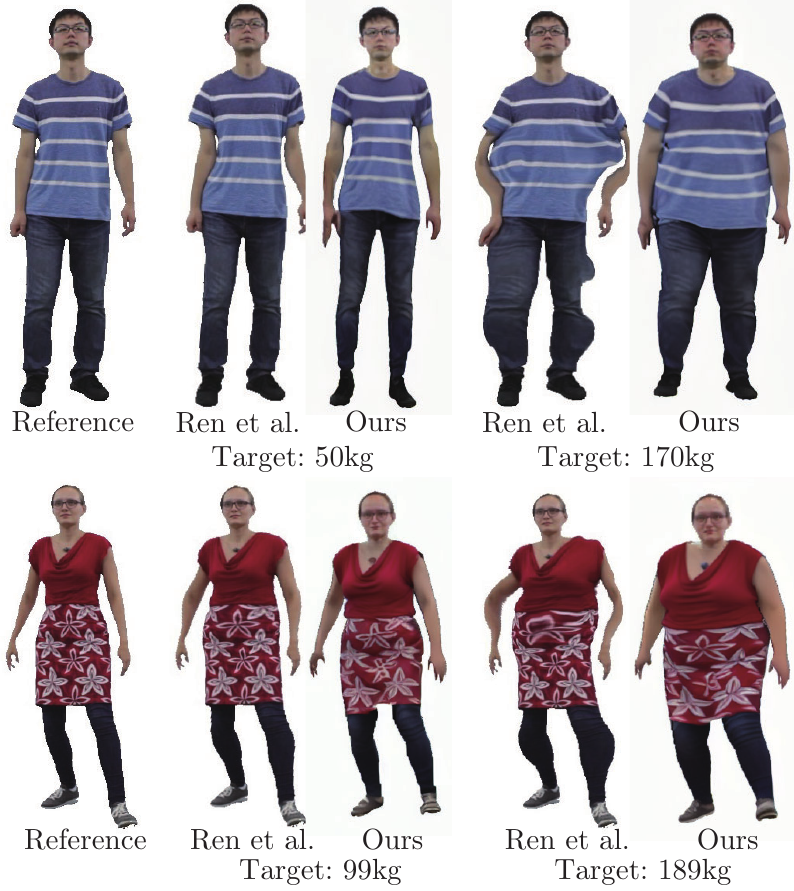}
 \caption{Qualitative comparison of editing the reference images to the target weights with our method and the method by Ren et al.~\cite{FlowBodyReshape}.}
 \label{fig:BodyShapeEdit}
\end{figure}

\subsection{Evaluation of body shape editing}
To the best of our knowledge, there is no dataset available for objective quantitative evaluation of body shape editing with large deformation. We therefore conducted a qualitative comparison between our method and the state-of-the-art method proposed by Ren et al.~\cite{FlowBodyReshape}. The existing method controls the body shape using a warping strength as input instead of a body height and weight. For fair comparison, we searched for an appropriate warping strength so that the silhouette of the output image aligns with that of the SMPL-X model with a body height and weight specified in our method.
\paragraph{Qualitative comparison.}Figure~\ref{fig:BodyShapeEdit} shows the qualitative results. In the results of the existing method, increasing body size often causes significant distortion in the torso regions. In contrast, our method can create plausible images. In addition, our method can handle changes in the facial appearance that occur with changes in body weight.



\subsection{Ablation studies}
\label{sec:abs}

\begin{figure}[t]
\centering
 \includegraphics[width=\linewidth]{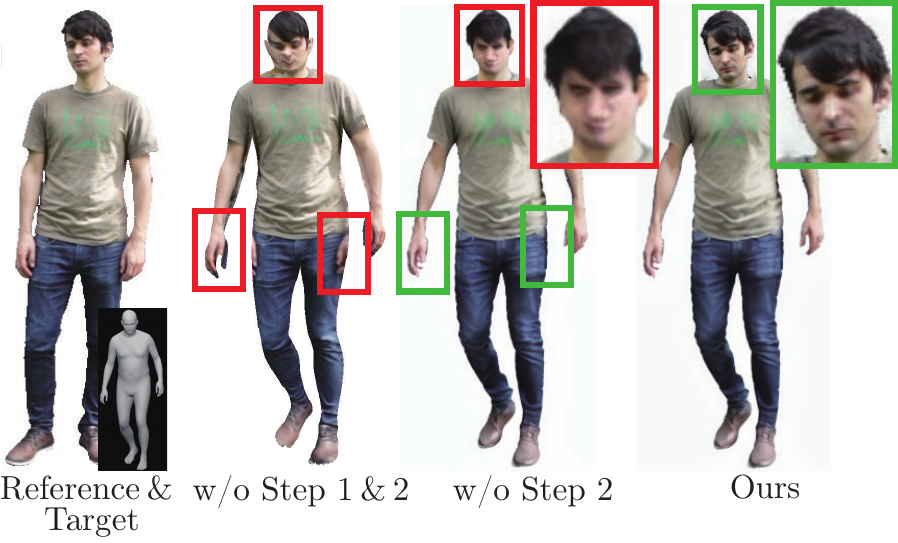}
 \caption{Qualitative results of the ablation study for diffusion-based refinement (\ykA{Steps~1 and 2}). As shown in the red boxes in the second column, without both \ykA{Steps~1 and 2}, we can observe not only artifacts on the hands and pants but also a distorted face. The third column shows that Step~1 removes these artifacts (green boxes), but the identity of the face is still not improved (red boxes). Step~2 can address this problem, as shown in the green boxes in the fourth column. }
 \label{fig:ablation_wo_step2}
\end{figure}

We conducted an ablation study on the MonoPerfCap dataset~\cite{MonoPerfCap} to evaluate the effectiveness of our diffusion-based image refinement (i.e., \ykA{Steps~1 and 2}). Table~\ref{table:ablation_wo_step2} and Figure~\ref{fig:ablation_wo_step2} show the quantitative and qualitative results, respectively. These results demonstrate that our fullbody refinement (Step~1) improves realism (FID) of the output images. This realism improvement also benefits the accuracy of pose estimation (Pose). However, we can see that facial identity (ID) degrades because the VAE used in the LDM cannot reconstruct relatively small faces accurately from low-dimensional latent maps. 
Our facial refinement (Step~2) can improve facial identity while preserving the scores of the other metrics. 
\begin{table}[t]
\caption{Quantitative results of the ablation study for diffusion-based refinement (\ykA{Steps~1 and 2}). }
\label{table:ablation_wo_step2}
\centering
\small
\vspace{0.25cm}
\begin{tabular}{l|c|c|c|c}
\hline
& LPIPS $\downarrow$ & FID $\downarrow$ & ID $\downarrow$ &  \yoA{AW} $\downarrow$ \\
\hline \hline
w/o \ykA{Steps~1 \& 2} & 0.242 & 55.862 & 0.232& 3.68\\
w/o Step~2 & \textbf{0.239} & \textbf{45.949} & 0.403 & \textbf{1.74} \\
Ours  & 0.242 & 46.432  &\textbf{0.175} & 1.82\\
\hline
\end{tabular}
\end{table}

\begin{figure}[t]
\centering
 \includegraphics[width=0.9\linewidth]{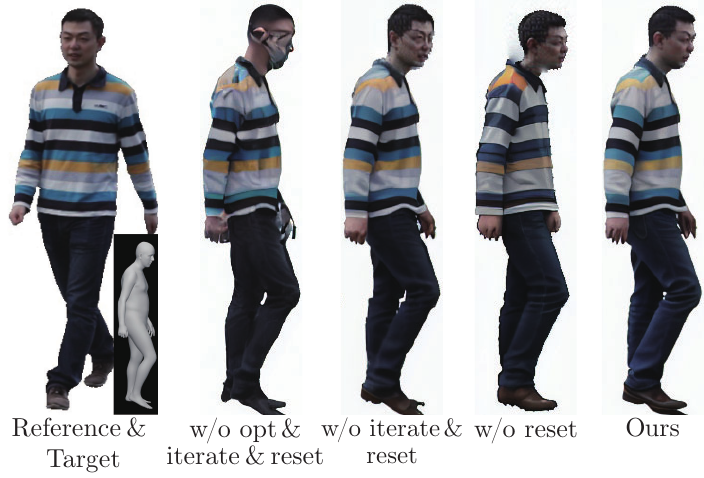}
  \caption{Qualitative results of the ablation study for our refinement approaches including text embedding optimization (opt), iterative refinement (iterate), and input reinitialization (reset).}
 \label{fig:ablation_iterative}
\end{figure}

In addition, we conducted another ablation study to evaluate our individual refinement approaches described in Section~\ref{sec:step1}. Table~\ref{table:ablation_iterative_refinement} and Figure~\ref{fig:ablation_iterative} show the quantitative and qualitative results, respectively. Here, ``opt'' means text embedding optimization, ``iterate'' means iterative refinement, and ``reset'' means input reinitialization. As shown in the comparison between ``w/o opt \& iterate \& reset'' and ``w/o \& iterate \& reset,'' text embedding optimization improves realism and facial identity. From the results of ``w/o reset'' and ours, we can also see that only iterative refinement does not perform well on its own but is effective when combined with input reinitialization. 
\begin{table}[t]
\caption{Quantitative results of the ablation study for our refinement approaches including text embedding optimization (opt), iterative refinement (iterate), and input reinitialization (reset).}
\label{table:ablation_iterative_refinement}
\centering
\footnotesize
\vspace{0.25cm}
\begin{tabular}{l|c|c|c|c}
\hline
  & LPIPS $\downarrow$ & FID $\downarrow$ & ID $\downarrow$ & \yoA{AW} $\downarrow$ \\
\hline \hline
w/o opt \&  iterate \& reset   & 0.244 & 56.093  & 0.404 &1.93 \\
w/o iterate \& reset   & \textbf{0.240} & 47.770   &0.183 &2.04\\
w/o reset    & 0.277 & 56.670   &0.187 &2.37\\
Ours & 0.242 & \textbf{46.432}  &\textbf{0.175} & \textbf{1.82}\\
\hline
\end{tabular}
\end{table}

\section{Conclusion}
In this paper, we proposed a method for editing the pose and body shape of \ykA{a} fullbody human \ykA{image}. Our method leverages \ykA{a} 3D parametric \ykA{body} model to control the pose and body shape using keypoints, height, and weight as input. Our two-step refinement pipeline is based on image-to-image translation with an LDM and refines the body and facial regions of the textured 3D human models. To improve the quality of the output images obtained via refinement, we introduced iterative refinement, text embedding optimization, and input reinitialization into the refinement pipeline. In the task of pose editing, our method exhibited quantitatively and qualitatively superior or comparable results compared to existing methods on multiple datasets. Our method also shows more plausible results in body shape editing. In the future, we will explore ways to speed up \ykA{our} method and enhance its capability to handle loose clothing like skirts.


{\small
\Urlmuskip=0mu plus 1mu\relax
\bibliographystyle{ieee_fullname}
\bibliography{arXiv}
}

\newpage


\appendix

\section{Implementation Details}
\subsection{\ykA{Initial textured 3D body construction}}
\begin{figure}[h]
\centering
 \includegraphics[width=\linewidth]{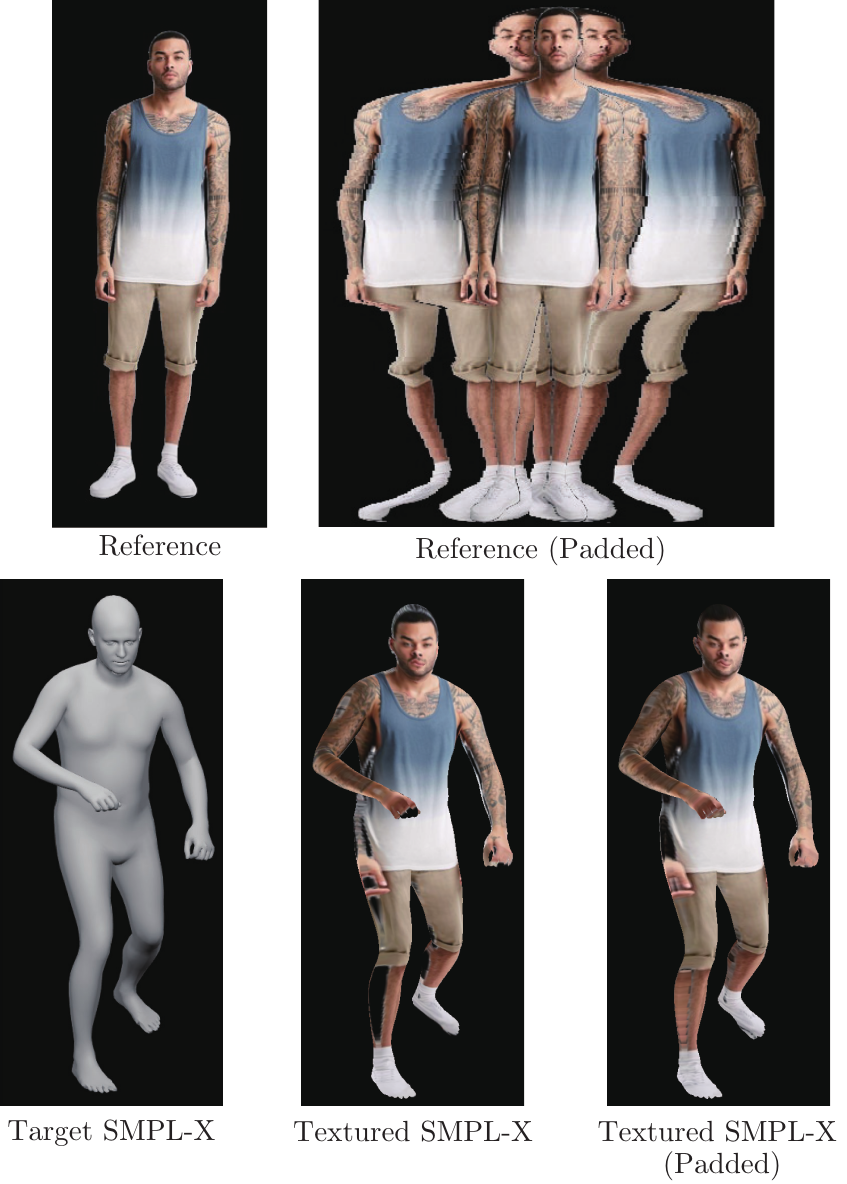}
   \caption{\ykA{Horizontal reflection} padding is applied to the reference image to prevent the background color from showing up in the texture-projected human model.}
 \label{fig:ProjectTexture}
\end{figure}

\paragraph{\ykA{Projective texture mapping.}}
\label{ProjectTexture}
We explain how to \ykA{construct an initial textured body model} with the desired pose and body shape.
First, \ykA{we fit} the SMPL-X model~\ykA{\cite{SMPL-X:2019}} to the reference image using \ykA{an} existing method~\cite{pymaf2021}.
For the reference person image and the fitted SMPL-X model, 
\ykA{we assign} UV coordinates \ykA{to corresponding vertices via projective} texture mapping. 
\ykA{We then change the pose and shape of the SMPL-X model to obtain an initial textured 3D body model.}
\paragraph{\ykA{Horizontal reflection padding.}}
\ykA{Na\"{i}vely applying projective texture mapping yields visual artifacts, particularly around the body's silhouette, due to slight misalignment. For example, the black background color appears around the right hand and right leg in the example of Figure~\ref{fig:ProjectTexture}, lower-middle. As a simple remedy for this, we apply horizontal reflection padding to the original reference image using a binary mask; for each scanline from slightly inside the mask, we copy pixel values at the mirror-symmetric positions about the mask boundary (Figure~\ref{fig:ProjectTexture}, upper-right).
This approach is not a perfect solution but is sufficient to avoid copying the background color (Figure~\ref{fig:ProjectTexture}, lower-right).}

\subsection{Loss functions}
\ykA{Here we describe the details of loss functions used in Steps~1 and 2.}

\paragraph{\ykA{Step~1:}~Fullbody refinement\ykA{.}}

\ykA{For the refinement of a fullbody image, we use the \yoA{Adaptive Wing (AW)} loss\cite{wingloss} $\mathcal{L}_{AW}$ and CLIP similarity~\cite{CLIP} loss $\mathcal{L}_{CLIP}$. The \yoA{AW} loss $\mathcal{L}_{AW}$ is the adaptive wing loss~\cite{wingloss} defined between the joint heatmaps estimated using OpenPose~\cite{OpenPose} for the output and rendered SMPL-X images. Our heatmap resolution is $128\times128$. CLIP similarity loss $\mathcal{L}_{CLIP}$ is defined} between the output and reference images for each part~\cite{elicit} based on SMPL-X labeling \ykA{as follows:}
\begin{equation}
    \mathcal{L}_{\text{CLIP}} = \sum_{p}^{l} \phi(I_{\text{ref}}^p)^{T} \phi(I_{\text{out}}^p),
\end{equation}
where $l$ is the number of body part \ykA{labels} of SMPL-X, $I_{ref}$ and $I_{out}$ \ykA{are} the body parts cropped \ykA{from the} reference and output image\ykA{s}. $\phi$ is the normalized embedding function of the CLIP.
\ykA{The total loss function for fullbody refinement is:}
\begin{equation}
    \mathcal{L}_{Fullbody} = \lambda_{\yoA{AW}} \mathcal{L}_{\yoA{AW}} +  \lambda_{CLIP} \mathcal{L}_{CLIP}\ykA{,}
\end{equation}
\ykA{where $\lambda_{Pose}=0.002$ and $\lambda_{CLIP}=2$ are the weights.}


\paragraph{\ykA{Step~2:}~Facial refinement\ykA{.}}
To optimize the text embedding for \ykA{refining a} face, we use
the identity loss \ykA{using} MagFace~\cite{meng2021magface}, \ykA{the} keypoint loss \ykA{using} RetinaFace~\cite{retinaface}, and \ykA{the} CLIP similarity~\cite{CLIP}. \yoA{The keypoint loss $\mathcal{L}_{keypoint}$ is defined as MSE loss between the face keypoints estimated using RefinaFace~\cite{retinaface} for the output and rendered SMPL-X images.} Unlike fullbody refinement, we simply measure the CLIP similarity between the reference and the output face image.
\ykA{The total loss function for the facial refinement is:}
\begin{equation}
    \mathcal{L}_{Face} = \lambda_{ID} \mathcal{L}_{ID} +  \lambda_{CLIP} \mathcal{L}_{CLIP} + \lambda_{\yoA{Keypoint}} \mathcal{L}_{\yoA{Keypoint}}\ykA{,}
\end{equation}
\ykA{where $\lambda_{\yoA{Keypoint}} = 0.1$, $\lambda_{CLIP} = 10$, and $\lambda_{ID} = 10$ are the weights. When we edit the body shape, we halve $\lambda_{CLIP}$ and $\lambda_{ID}$ to tolerate changes in facial features.}

\subsection{Text prompt}
\yoA{We describe the details of prompts used for conditioning on our refinement module. Our prompts contain ``sks,'' a special token used for text-to-image personalization by DreamBooth~\cite{ruiz2022dreambooth}. Our method associates this token with a reference person. In addition, our prompts contain information on a target face orientation, such as ``facing left.'' We used the face detection API of Face++~\cite{facepp} to obtain the face orientation, which is automatically reflected in the prompts. For body shape editing, we use adjectives describing the body shape according to BMI calculated from the input height and weight (see Table~\ref{tab:adj_and_bmi}). For example, when the target model faces to the left with a fat body, we use a prompt ``photo of a fat sks man facing left'' \yoA{in Step~1. In Step~2, we use ``face'' instead of ``man''}. 
}

\begin{table}[t]
\caption{Adjectives describing body shape corresponding to BMI.}\centering
\begin{tabular}{c|c}
\hline
\centering
 BMI&  adjective \\
\hline \hline
$\leq$ 15.0 & ``skinny" \\
$\leq$ 18.5 & ``under weight" \\
$\leq$ 25.0 &  \\
$\leq$ 30.0 & ``overweight" \\
$\yoA{>}$ 30.0  & ``fat" \\
\hline
\end{tabular}
\label{tab:adj_and_bmi}
\end{table}

\subsection{Refinement mask}
\yoA{
We describe how to create a refinement mask, which indicates areas to be refined. In Step~1, we compute a mask consisting of invisible areas in a reference person image. To do so, we first emit a ray to each triangle's centroid in a SMPL-X~\cite{SMPL-X:2019} mesh from the viewpoint for texture projection. Next, we assign an ``invisible'' label to the triangles that the rays do not hit. After editing the pose and body shape of the SMPL-X model, we render the edited model to obtain a mask according to the labeled areas. In Step~2, we compute a mask by assigning 0 to pixels within 20\% of the mask width from its boundaries, measured using the Manhattan distance, and 1 to the remaining pixels. }

\section{Additional Result\ykA{s}}
\ykA{We show the additional results that are not included in the main paper due to the page limitation. The reference images of the following results were obtained from} DeepFashion~\cite{DeepFashion}, MonoPerfCap~\cite{MonoPerfCap}, Everybody Dance Now~\cite{everybodyDanceNow}\ykA{,} and EHF~\cite{SMPL-X:2019}.
\subsection{Qualitative evaluation}
\subsubsection{Evaluation of body shape editing}
\begin{figure*}[t]
\centering
 \includegraphics[width=\linewidth]{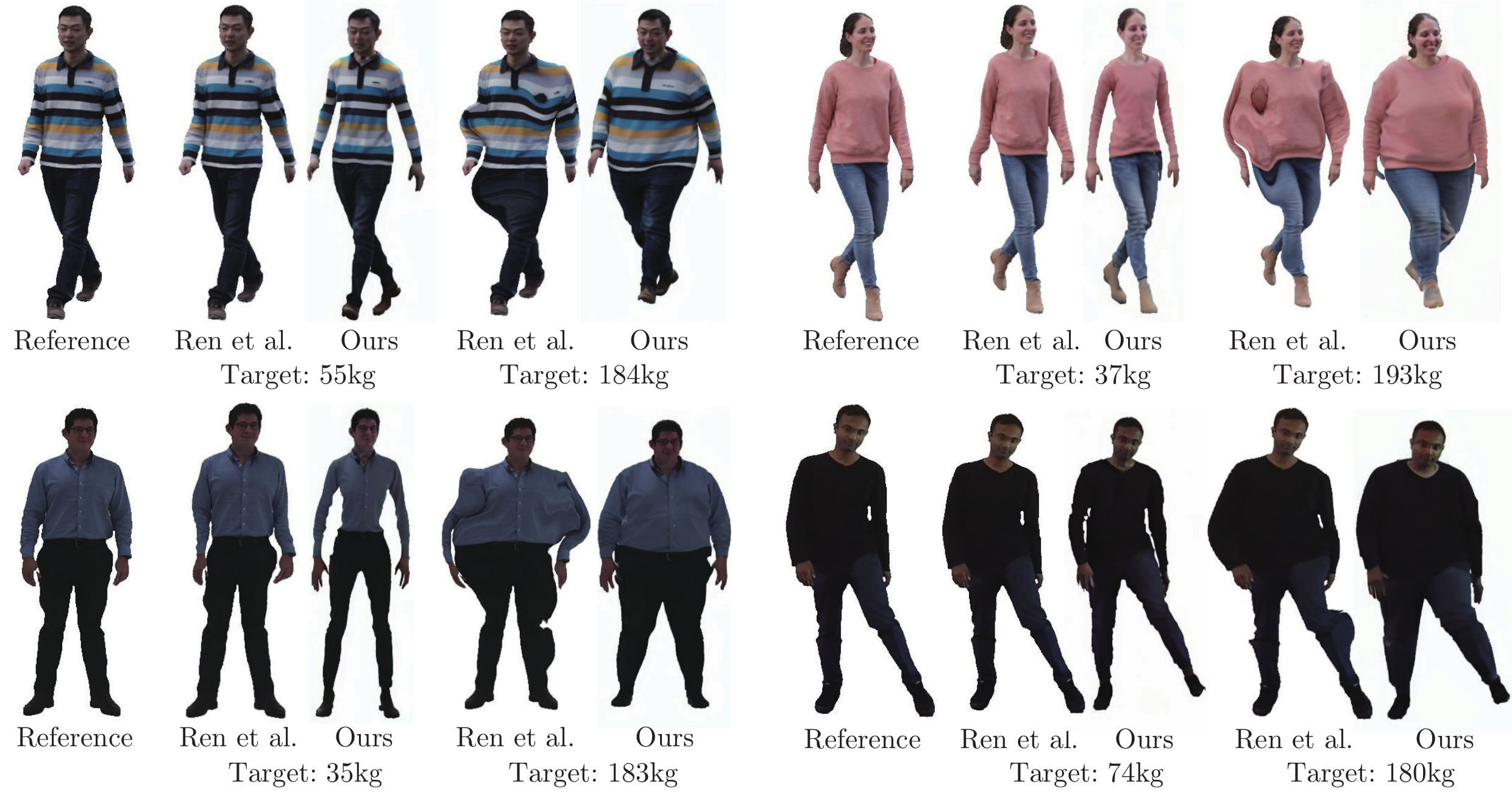}
 \caption{Qualitative comparison of editing the reference images to the target weights with \ykA{the method by} Ren et al.~\cite{FlowBodyReshape} \ykA{and ours}.}
 \label{fig:body_shape_editing_results}
\end{figure*}

We conducted a qualitative comparison with the state-of-the-art body shape editing method by Ren et al\ykA{.}~\cite{FlowBodyReshape} in the same way \ykA{as their} paper.
\ykA{Figure~\ref{fig:body_shape_editing_results} shows the results.}
In the results of \ykA{their} method, increasing \ykA{the} body size often causes significant distortion in the torso. In contrast, our method can create plausible images. \ykA{Our} method can \ykA{also} handle \ykA{facial appearance changes} that occur \ykA{along with the body weight changes}.

\subsubsection{Evaluation of pose and body shape editing} 
\begin{figure*}[t]
\centering
 \includegraphics[width=\linewidth]{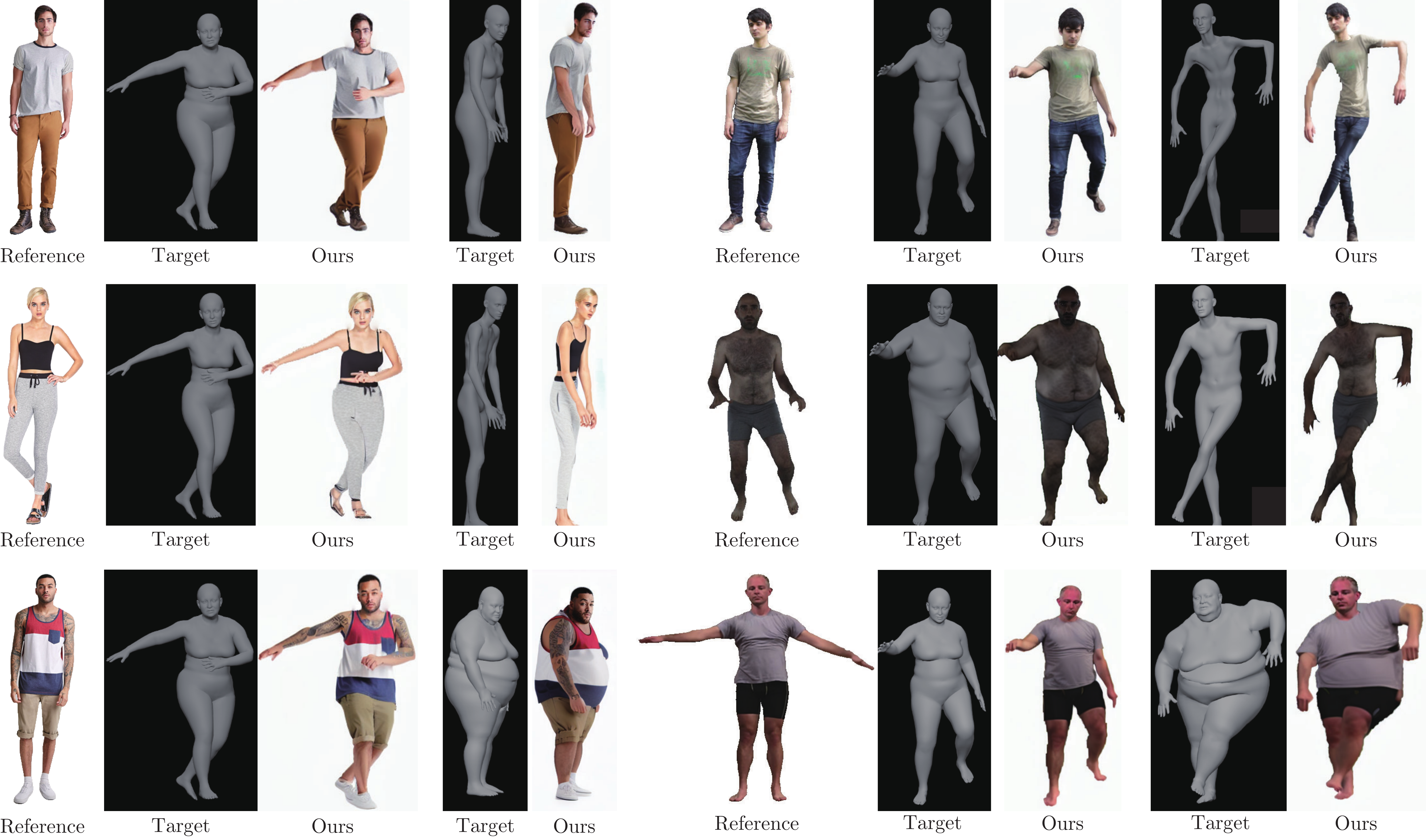}
 \caption{Qualitative comparison of \ykA{simultaneous edits of both} target's \ykA{poses} and body \ykA{shapes} in reference images using our method. The \ykA{edited results are} plausible \ykA{with identity preservation} in terms of clothing and facial features.}
 \label{fig:pose_and_body_shape_editing_results}
\end{figure*}

\ykA{Figure~\ref{fig:pose_and_body_shape_editing_results} shows our unprecedented results in which both poses and body shapes were edited at the same time. Such simultaneous edits have been infeasible with existing methods, to the best of our knowledge. The results demonstrate that our method can edit} the target pose and body shape \ykA{simultaneously} while maintaining the subject's identity in terms of clothing and facial features.

\subsection{Ablation Study}
\subsubsection{Facial degradation \ykA{by VAE}}
\begin{figure}[t]
\centering
 \includegraphics[width=\linewidth]{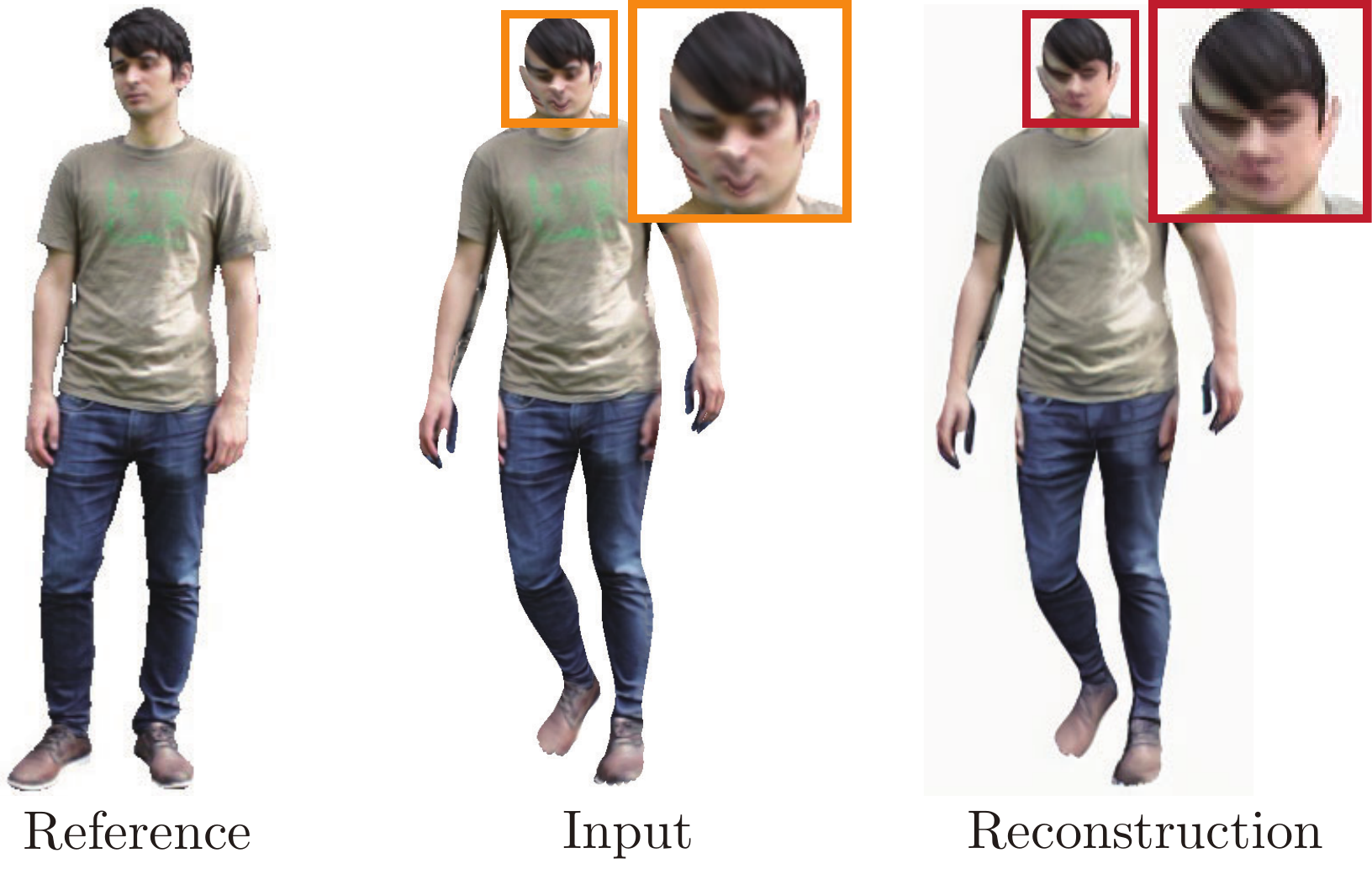}
  \caption{Qualitative comparison of decreased facial quality when simply reconstructing images using VAE.}
 \label{fig:degrade_vae}
\end{figure}

In the LDM used in our method, the \ykA{face} quality is degraded by simply reconstructing the input image with VAE. This is because the VAE used in the LDM cannot reconstruct relatively small faces accurately from low-dimensional latent maps. An example of the degradation is shown in Figure~\ref{fig:degrade_vae}, and the quantitative evaluation metrics are shown in Table~\ref{tab:metrics_vae_degrade}. 
\ykA{These results warrant our approach that extracts a face region and refines it separately.}

\begin{table}[t]
\caption{Quantitative evaluation when images are simply reconstructed with VAE.}\centering
\begin{tabular}{l|c|c|c|c}
\hline
 & SSIM $\uparrow$  & LPIPS $\downarrow$  & FID $\downarrow$& ID $\downarrow$\\
\hline \hline
Input & 0.714 & 0.243 & \textbf{55.862} & \textbf{0.232} \\
Reconstruction & \textbf{0.716} &  \textbf{0.242}  &  65.484 & 0.375 \\
\hline
\end{tabular}
\label{tab:metrics_vae_degrade}
\end{table}

\subsubsection{Refinement with weak noise}
\begin{figure}[t]
\centering
 \includegraphics[width=\linewidth]{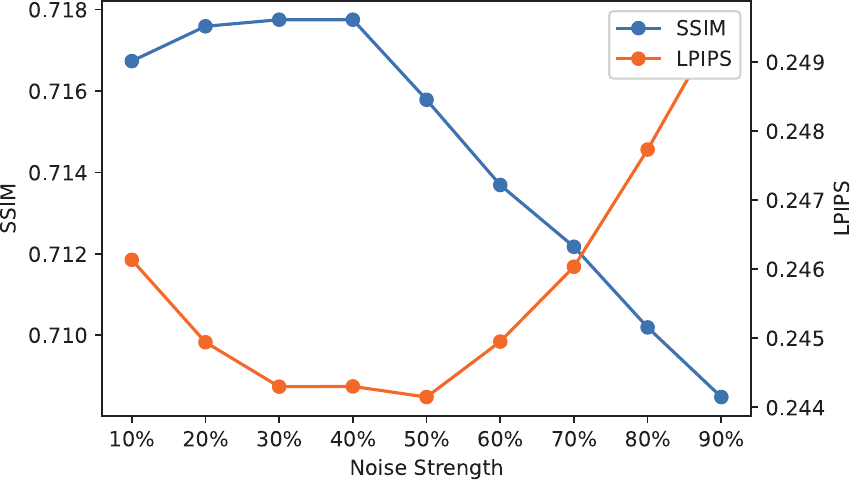}
 \caption{\ykA{Graphs} depicting the variation\ykA{s of} SSIM and LPIPS scores \ykA{with varying noise intensities}.}
 \label{fig:ablation_noise_strength}
\end{figure}

\ykA{To find an appropriate noise intensity for our iterative refinement, we experimented with single iterations of refinement with different noise levels. We increased the noise level from 10\% to 90\% in increments of 10 percentage points. Table~\ref{tab:ablation_noise_strength} summarizes the qualitative evaluation, and Figure~\ref{fig:ablation_noise_strength} shows graphs of SSIM and LPIPS with varying noise intensities. These results revealed that weaker noise tends to yield better results regarding pixel-level metrics such as PSNR and SSIM}
because weaker noise preserves the projected textures as they are. On the other hand, for more perceptual metrics such as LPIPS and FID, the values tend to be optimal around 30\% to 50\% noise intensity, with performance degrading as the noise intensity deviates from this range. This pattern suggests that\ykA{,} around 30\% to 40\% noise intensity, \ykA{we} can effectively correct unnatural areas while preserving the texture of the input image. 
\ykA{Consequently, we conclude that weak noise intensities from 30\% to 40\% seem effective for refinement.}
In our approach, we choose 30\% noise intensity to balance computational efficiency while maintaining effectiveness.


\begin{table}[t]
\caption{Quantitative evaluation metrics for a single \ykA{r}efinement with \ykA{varying noise intensities}.}
\centering
\vspace{0.25cm}
\begin{tabular}{c|c|c|c|c}
\hline
 & PSNR $\uparrow$  & SSIM $\uparrow$  & LPIPS $\downarrow$& FID $\downarrow$\\
\hline \hline
10\% & 19.613 & 0.717 & 0.246 & 59.221 \\
20\% & 19.650 & \textbf{0.718} & 0.245 & 57.808 \\
30\% & \textbf{19.669} & \textbf{0.718} & \textbf{0.244} & 56.093 \\
40\% & \textbf{19.669} & \textbf{0.718} & \textbf{0.244} & 56.084 \\
50\% & 19.561 & 0.716 & 0.244 & \textbf{50.748} \\
60\% & 19.582 & 0.714 & 0.245 & 52.582 \\
70\% & 19.519 & 0.712 & 0.246 & 51.816 \\
80\% & 19.433 & 0.710 & 0.248 & 51.592 \\
90\% & 19.356 & 0.708 & 0.250 & 52.011 \\
\hline
\end{tabular}
\label{tab:ablation_noise_strength}
\end{table}

\end{document}